\documentclass[nonacm,sigconf,10pt]{acmart}

\usepackage[caption=false,font=footnotesize,labelfont=sf,textfont=sf]{subfig}
\usepackage{pifont}
\usepackage{xspace}

\renewcommand\footnotetextcopyrightpermission[1]{} % removes footnote with conference information in first column
\definecolor{darkgreen}{rgb}{0.078,0.667,0.016}

\newcommand{\outline}[1]{}
\newcommand{\Eg}{\textit{E.g.}\xspace}

\newcommand{\ie}{\textit{i.e.}\xspace}

\usepackage{grumble}
\AtBeginDocument{%
  \providecommand\BibTeX{{%
    \normalfont B\kern-0.5em{\scshape i\kern-0.25em b}\kern-0.8em\TeX}}}

\begin{document}

%%
%% The "title" command has an optional parameter,
%% allowing the author to define a "short title" to be used in page headers.
%%XM \title[Balancing Resources between Data Preprocessing and Training ...]{Balancing Resources between Data Preprocessing and Training for Effective End-to-End Training of Deep Neural Networks}

\title{Understand Data Preprocessing for Effective End-to-End Training of Deep Neural Networks}

\author{\fontsize{10}{15}Ping Gong$^{1}$, Yuxin Ma$^{1}$, Cheng Li$^{1,4}$, Xiaosong Ma$^{2}$, Sam H. Noh$^{3}$}

\affiliation{%
  \institution{$^{1}$University of Science and Technology of China \quad $^{2}$Qatar Computing Research Institute \quad $^{3}$Virginia Tech\\$^{4}$Anhui Province Key Laboratory of High Performance Computing}
  %\streetaddress{No. 96, Jinzhai Road}
  %\city{Hefei}
  %\state{Anhui}
  %\postcode{270000}
  \country{}
%  \vspace{0.5ex}
}
%%
%% The abstract is a short summary of the work to be presented in the
%% article.
\begin{abstract}

In this paper, we primarily focus on understanding the data preprocessing pipeline for DNN Training in the public cloud. First, we run experiments to test the performance implications of the two major data preprocessing methods using either raw data or record files. The preliminary results show that data preprocessing is a clear bottleneck, even with the most efficient software and hardware configuration enabled by NVIDIA DALI, a high-optimized data preprocessing library. Second, we identify the potential causes, exercise a variety of optimization methods, and present their pros and cons. We hope this work will shed light on the new co-design of 
``data storage, loading pipeline'' and ``training framework'' and flexible resource configurations between them
%or three items: ``data storage,'' ``loading pipeline'' and ``training framework''???}
%data storage, loading pipeline and training framework, 
so that the resources can be fully exploited and performance can be maximized.
\end{abstract}

%%
%% This command processes the author and affiliation and title
%% information and builds the first part of the formatted document.
\maketitle

\section{Introduction}
\label{sect:intro}

%\sam{Here is another take on the title: (Position) The Need to Consider Data Loading for Effective End-to-End Training of Deep Neural Networks}

%\sam{Another global matter: The names of frameworks need to be unified to its original use, e.g.,: MxNet $==>$ MXNet, Tensorflow $==>$ TensorFlow, Pytorch $==>$ PyTorch}\cheng{I will perform all changes.}
%% DNN popular and DNN training is computational intensive,
%% using GPUs

%%XM Deep learning is now widely used in many areas, such as image recognition, speech recognition, recommendation systems, search engines, etc~\cite{angelova2015pedestrian, ba2014multiple, frome2013devise, gonzalez2015frame}. 
One major contributor to the huge success of deep learning (DL) is the existence of diverse and large datasets, while another is the abundance of massive computing power, mostly from accelerators like GPUs.
%%XM Recently, to meet the demand of big data analysis over fast-growing datasets, many 
Popular DL frameworks, such as MXNet~\cite{MXNet}, TensorFlow~\cite{TensorFlow}, and PyTorch~\cite{Pytorch}, connect these two driving forces to train deep neural network (DNN) models,
%%XM using a massive number of accelerators like GPUs.\ The training process works iteratively by 
iteratively feeding GPUs with batches of data samples from a collection of raw data 
%%XM , which are transformed 
such as images, videos, and text. 
Before being efficiently consumed by armies of GPU cores, raw data must first be loaded into memory and processed through a sequence of transformation operators such as \texttt{decode}, \texttt{crop}, \texttt{resize}, and \texttt{normalize}.%\footnote{One example of such an end-to-end training script is shown in ~\cite{}.}

%% We observe that there exist performance bottlenecks
%% regarding data loading pipeline
To effectively utilize the massive GPU parallelism available today, such data preprocessing must also be effective enough to continually pump data to satisfy the GPUs' appetite.
%%XM may consume a large amount of resources to feed the GPUs with enough data samples. Thus, its performance is a crucial factor in determining the speed and resource efficiency of end-to-end model training. 
Unfortunately, most existing studies have focused on just the training part, often neglecting the data preprocessing pipeline~\cite{Goyal2017AccurateLM, li2019a,Koliousis2019Crossbow}. 
Fortunately, though a few recent studies have started to consider this issue.
\Eg, Uber and Apple introduced their data platforms, 
%%XM which mostly concentrate on
with data models and methods to store and manage data for machine learning tasks~\cite{apple.mldb.10.1145/3299869.3314050,uber.petastorm}. 
Yang and Gong proposed a distributed cache to speed up the I/O steps in data preparation for distributed DNN training~\cite{Yang2019CacheLoad}, 
%\sam{reference [30] looks to be incomplete; 
%\sam{for references [10] and [13], the full name needs to be listed, instead of abbreviations.}\cheng{done}
while OneAccess applies a similar idea to avoid redundant data loading for concurrent hyper-parameter tuning experiments sharing
%%XM that use the same 
datasets~\cite{OneAccess}. 
Still, none of the existing studies provide a comprehensive overview of DL data preprocessing pipeline, its performance, or major systems design issues involved. 

%% We conduct a study and report results, summary
This work examines the data preprocessing pipeline in three major frameworks, specifically, TensorFlow, MXNet, and PyTorch, and its performance implications on overall training efficiency. 
Our major findings are as follows:
% \vspace{-5pt}
\begin{itemize}
    \item The widely used DL frameworks have long and highly similar data preprocessing pipelines, with the common data format and data structure choices.
    %\item There are some common design choices. For instance, iterable datasets, composable data loading pipeline, degree of parallelism, caching, shuffling, ..., etc.
    \item Different DNN models, meanwhile, show different relative speeds in data preprocessing and actual training. 
    With most studies focused on training performance, the majority of model training workflows have bottlenecks at data loading and preprocessing.  
    %%XM \item Clearly, the data loading is a performance bottleneck, which impedes the overall training process. The degradation ranges from xx\% to xx\%. The data loading is computation, memory, I/O intensive. \cheng{Please check the last two.}
    %\item Offline pre-processing has limited benefit, with intermediate results rarely shared across many jobs. \xm{We need to be assertive here, otherwise the whole investigation would go down this path.}
    \item 
    While the aforementioned three major frameworks all perform data preprocessing on CPUs and training on GPUs, the new NVIDIA framework DALI~\cite{DALI} does deliver better overall performance by enabling CPU-GPU co-processing for preparing training data samples.
    %%XM , compared to the built-in preprocessing libraries in the three frameworks. 
    However, it possesses caveats such as potential Out-Of-Memory (OOM) problems, currently requiring a manual search for a proper batch size to avoid OOM while preserving high throughput.
    %\textcolor{pink}{(with certain caveats such as potential Out-Of-Memory (OOM) problems). This would require some manual work to find the proper batch size for avoiding OOM while preserving high throughput.}
    \item Nevertheless, a single time-consuming step (such as image decoding, currently placed at the CPU side by DALI) may dominate the pipeline execution time, making GPUs wait for data. To eliminate this bottleneck, one would need to add more CPU resources.
    \item %\sam{
%%    While some cloud providers are starting to provide flexibility in purchasing resource instances (See Table~\ref{tab:cloudsurvey}), others still only offer static instance settings with synchronized scaling of both CPUs and GPUs leaving one resource inevitably under-utilized (See Figure~\ref{fig:res_utilization}). 
    Our preliminary results indicate that more flexible node configurations would allow more cost-effective executions of diverse DL pipelines:
    7.86\% and 3.03\% speedup of training speed for AlexNet and ResNet50 with more CPU resources allocated, respectively, and 75\% reduction in CPU resource allocation for ResNet50 with relatively comparable performance. Loading data samples from SSD-based storage options almost delivers the best training speed for ResNet18, while using DRAM introduces a 1.84$\times$ speedup for AlexNet.
%%    However, even with the flexibility provided, it is not easy to find the best configuration for one's model of choice.
%%    A user-friendly tool to find that configuration becomes a necessity.
\end{itemize}   
    
    %%XM Also, while providers such as Google Cloud has started to allow some degree of configuration flexibility, finding the right resource configuration for the model at hand is not a easy task.
    %}
    
    %\textcolor{green}{really strange (and ignore this line). I get a compile error if I remove this line:: \sout{whether to say this or not: Our eventual goal is provide a tool to}}
    
    %\item \textcolor{pink}{Unfortunately, public cloud providers often offer instances with synchronized scaling up in both CPUs and GPUs, leaving one side inevitably under-utilized. Our preliminary results indicate that more flexible node configuration would allow more cost-effective executions of diverse DL pipelines. \cheng{But Google Cloud allows some degree of flexible configuration.}} 
    %%XM We also observe that different stages on the data loading pipeline have highly-varied time costs and resource demands. Decoding is the most time-consuming and resource-hungry part.
    %%XM \item GPU can improve the performance of data loading, but also result in resource contention, which lead to OOM problems.

\begin{figure*}[t!]
  \centering
  \includegraphics[scale=0.35]{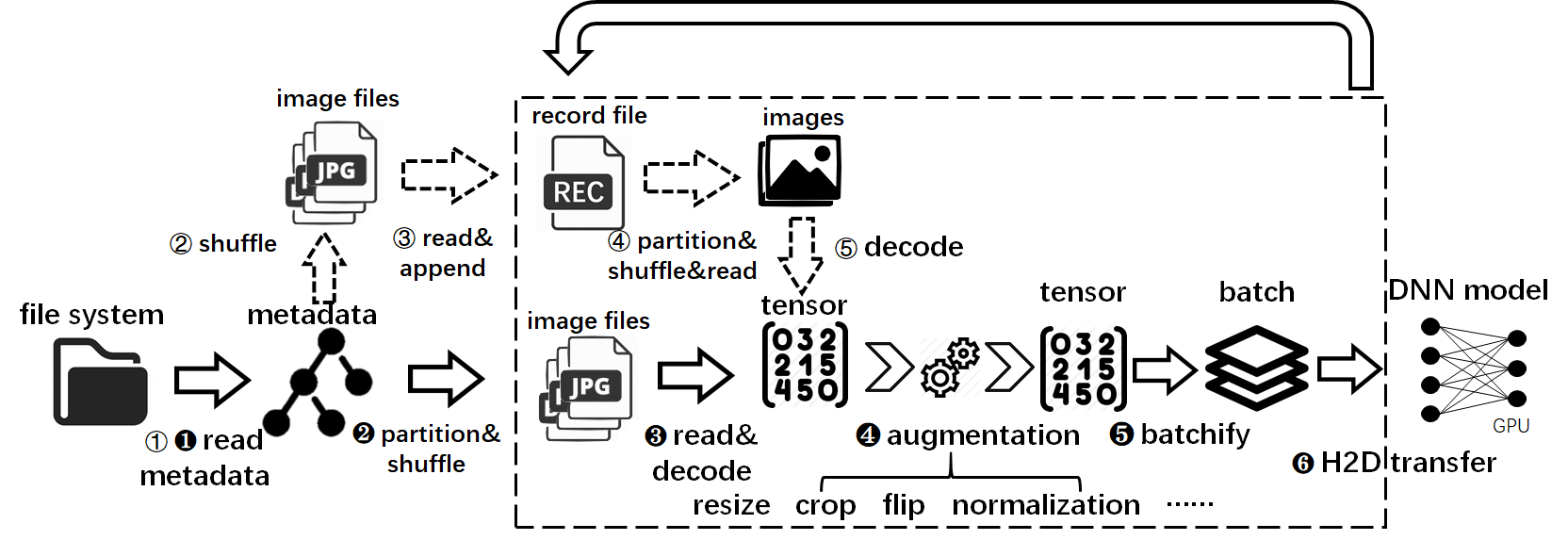} 
  \caption{The end-to-end training pipeline where the first stage is the data preprocessing for Computer Vision tasks.
  } 
  \label{fig:pipeline}
\end{figure*}

Our study chimes in the argument for \textit{resource disaggregation}~\cite{han2013network,klimovic2016flash,Gao2016NetDis,Dedrick2019Disaggregation}, a new trend in cloud infrastructure design featuring
%%XM Compared to traditional monolithic deployments, the 
composable and disaggregated resource deployment, %\cred{
delivering comparable performance, while retaining strong flexibility at the same time~\cite{Xu2017Analysis, Shan2018LogOS}. %}
Currently, some cloud providers such as Google Cloud and EC2 are enabling this feature to some extent, by allowing the number of vCPUs for GPU machines to be adjusted, though under particular constraints~\cite{GoogleGPUInstances}. 
We believe that the data preprocessing-training resource balancing problem described in this paper lends a common and resource-intensive use case to 
upcoming cloud resource disaggregation offerings. 
In particular, to avoid time-consuming manual configuration, there should be tools making such configuration and scaling decisions automatic, for higher overall user DL model training throughput (per hour or dollar) and simultaneously, for better overall system utilization from the cloud provider's point of view.

\section{Understanding the Training Pipeline}
\label{sect:backandmotiv}
%\sam{if space allows, say something here}
It is necessary to train DNN models with vast amounts of data to achieve accurate results. 
However, the real-world data sets are often not in a suitable form such that they can be directly fed into the neural network.
The data set often goes through a data preprocessing pipeline, spanning across loading, decoding, decompression, data augmentation, format conversion, and resizing.
Given that the performance of accelerators, including GPUs, is rapidly improving, this data loading pipeline phase of end-to-end training will inevitably become more and more significant. %In this section, we provide a discussion of end-to-end training with an in-depth focus on the data preprocessing pipeline.

\subsection{Iterative DNN training}
Training DNN models are computationally intensive and time-consuming. 
To shorten the training process and make DNN models quickly deployable, current common practice explores the massive parallelism offered by multiple GPUs for collective training following a mini-batch stochastic gradient descent (SGD) algorithm~\cite{Li2014MinibatchSGD}.
This phase of training, which is depicted (and denoted ``DNN model'') on the rightmost side of Figure~\ref{fig:pipeline}, has been the focus of the majority of studies on improving DNN training.
This training process works iteratively: at each round, a small batch of data samples will be fetched and loaded into the GPU memory as input for the neural network model training. 
However, for such training, typically, the application-specific raw data stored in persistent storage need to be transformed into \textit{tensors}, a unified data format consumed by various training frameworks.
We refer to this phase of training as a data preprocessing pipeline, which is depicted in the left part of Figure~\ref{fig:pipeline}. %In the rest of this section, we give a detailed description of this phase of training.

\subsection{Overview of data preprocessing pipeline}

We analyze four different implementations of the data preprocessing pipeline, where the first three are built-in solutions for three widely adopted training frameworks, namely, TensorFlow, MXNet, and PyTorch, and the last one is a unified interface for all three aforementioned frameworks developed by NVIDIA called DALI~\cite{DALI}. Despite being implemented by different communities, the composition of their pipelines looks almost the same. As follows, we report two major data preprocessing methods, namely, loading from raw data and loading from record files. 
%\textcolor{green}{
For simplicity, in this paper, we describe the image processing scenario as an example. %%s and describe the scenario. % using PyTorch terminology.
%}
%\sam{
However, we observe that the general scenario holds across tasks ranging from Natural Language Processing (NLP) to Video Processing. %\textcolor{pink}{??? need to verify}}

%%%\sam{Need to specify the model that is being considered in Figure 1. Also, why we consider the three frameworks; essentially that they are the most popular frameworks used today.} \cheng{Comment 1: Ping and Yuxin told me that this pipeline is general across all CNN models. Comment 2: addressed.}

\subsubsection{Raw data preprocessing}

%%\sam{I don't see the match between Figure 1 and the following description, especially the first part of the paragraph.} \cheng{Please check the revised figure.}

%\sam{Can we add numbers to the figure and match the numbers to the description given below? The numbering I use below may need to be adjusted accordingly.}

%\sam{
%\sam{Need to describe what \texttt{Data Loader} is.}
The numbers in the black circle in Figure~\ref{fig:pipeline} depict the preprocessing pipeline for raw data.
\ding{182} First, \texttt{Data Preprocessor} (or short, \texttt{DPP}) reads a sequential metadata file, and stores it in memory as a dictionary for retrieving the actual data later on. 
The metadata file is usually generated offline, and each of its tuples contains the index (sequentially assigned), label, and path of each image. %This step is called ``metadata reading''. 
Then, \ding{183} \texttt{DPP} partitions the whole file identifier list into a set of smaller sequences and shuffles them. %This step is called ``partition \& shuffle''. 
\ding{184} \texttt{DPP} continues to get the file ids from each sequence, uses ids to look up the corresponding labels and paths in the memory dictionary, loads the raw image files pointed by the paths, and decodes them into tensors. %This step is called ``read \& decode''.
%}
%%\textcolor{pink}{As shown in Figure~\ref{fig:pipeline}, the raw data loading pipeline can be divided into several major steps. To make this pipeline run, we need to first generate a sequential metadata file, in which each tuple contains the index (sequentially assigned), label and path of each image. There tuples are later read for further processing and retrieval. Once this offline generated metadata file is in place, \texttt{Data Loader} stores it in memory as a dictionary with id as key, and uses id as a sequence. This step is called ``metadata reading''. Then, \texttt{Data Loader} partitions the whole file identifier list into a set of smaller sequences and shuffles them. This step is called ``shuffle'' Following that, \texttt{Data Loader} gets file ids from each sequence, uses ids to look up the corresponding labels and paths in the memory dictionary, loads the raw image files pointed by paths, and decodes them into tensors. This step is called ``read and decode''.}
%\sam{
When tensors of the raw image files are ready, \ding{185} \texttt{DPP} moves to the ``augmentation'' step, where tensors are transformed into new ones via a sequence of steps such as random crop, flip, rotation, etc. 
Finally, \ding{186} \texttt{Data Loader} combines multiple tensors to form batches and \ding{187} load these batches to the GPUs. %These two final steps are called ``batchify'' and ``H2D transfer'', respectively.
%}

\subsubsection{Record file preprocessing}
%\sam{I tried rewriting the first part of the paragraph:}

As indicated in Figure~\ref{fig:pipeline} by the white circled numbers, in addition to raw data preprocessing, there is another method, which makes use of record files. 
These record files are generated, through offline processing, \ding{172} - \ding{174} by reading a large number of raw image files stored on disks and then, appending them into a few large sequential record files. 
These files include images as well as their label and id. 
%\textcolor{green}{
Then, during runtime, these record files are \ding{175} read into memory and partitioned into smaller chunks. 
These chunks are \ding{176} decoded, and thereafter, the remaining steps (\ding{185} - \ding{187}) are the same as the raw data preprocessing method. 
%}
%%\textcolor{green}{Then, during runtime, these record files are \ding{175} read into memory and partitioned into smaller chunks. After this step, all the remaining steps (\ding{184} - \ding{187}) are exactly the same as the raw data loading method. }
%%\sam{\ding{176} is missing in the description and saying ``all the remaining steps (\ding{184} - \ding{187})'' misses on \ding{184}.}
%%}

%\sout{As indicated in Figure~\ref{fig:pipeline}, in addition to the raw data loading, there is another loading method using record files. 
%A record file consists of many records, each of which includes an image plus its label and id. 
%The record file loading method works as follows: 
%First, we need to read a massive number of raw image files storing on disks and append them into a few large sequential record files at offline. Second, at runtime, record files are read into memory and partitioned into smaller chunks. 
%After this step, all the remaining steps are exactly the same as the raw data loading method. 
%}

There are two major reasons for designing this record file loading method. First, compared to raw data, record files transform random accesses into sequential accesses, which can lead to performance gains if data are stored in slow storage or scattered across multiple machines. 
%\sam{
Second, it is possible to offload some operators, such as crop and resize, in the data preprocessing pipeline to the record file generation phase for improved data loading efficiency.
%}
In essence, the record file loading method is trading extra processing time and storage space for training speed improvement.

%\sam{
Finally, one would think it feasible to take all tasks along the data preprocessing pipeline offline so that the runtime can be optimized. 
While this is feasible in terms of performance, this may not be the best approach for DNNs.
This is because the training process requires some form of randomness from data sample inputs to improve the robustness and universality of DNN models, which is achieved by executing some non-deterministic operators in the pipeline such as random rotates and flips.
Such randomness is required for the best results in DNN. Therefore, the preferable choice is to just append images to sequential record files with no further preprocessing. 
%}
%\textcolor{pink}{
%One would expect to perform all tasks alongside the data loading pipeline at offline so that the runtime can be fastest. However, it is \sout{impossible} available but may not get the best results. This is because the training process requires some form of randomness from data sample inputs to improve the robustness and universality of DNN model, which is achieved by executing some non-deterministic operators in the pipeline, such as random \sout{crop} rotate and flip, etc\cheng{Ping and Yuxin, please check} \ping{ done, but please recheck}. \sout{Those non-deterministic operators produce different results across different runs even with the same inputs, and thus cannot be just performed once. Besides those operators, we identify quite a few deterministic operators, such as decode, resize, etc} \cheng{@Ping and Yuxin, please check}. \ping{ have some expression problems, i will rewrite it}}

% \vspace{-5mm}
%%\subsection{Summary of mainstream frameworks}
\subsubsection{CPU-GPU co-processing}
%%\sam{It might be good to have a table that refers to the steps described above along with the associated commands used for each of the frameworks and DALI.}

%\sam{A short description of DALI here}
All the four data loading pipeline implementations that we consider allow both raw data and record file loading methods. %They also allow users to tune parameters to improve performance, such as batch\_size, parallelism, cache, etc\cheng{@Ping, Yuxin, could you please complete it}. 
The main difference between the three built-in data loading libraries in the TensorFlow, MXNet, and PyTorch frameworks and DALI is that while the former three-run the data loading pipeline on CPUs only, DALI makes use of the GPUs to speed up the task by offloading parts of the pipeline to GPUs. We refer readers to DALI code examples~\cite{DALICodeExample}, where the CPU-GPU co-processing can be configured manually by setting the device option to support data preprocessing operators.
%\sam{
This approach has the advantage of improving the data loading pipeline, but it has the disadvantage of making use of GPU resources. We experience out-of-memory (OOM) errors on GPUs when training ResNet18 with a 512 batch size and the FP32 precision, and eliminate these errors by reducing the batch size to 384.
%\textcolor{green}{
Enabling this feature could possibly force a reduction of the batch size to allow the sharing of GPU memory with the DNN training phase.
%}
%This can result in suboptimal training 
%\textcolor{green}{
%results~\cite{}.
%}
Furthermore, such adjustments also need to be done manually, which can be an arduous journey. 
\section{End-to-End Training Performance Analysis}
\label{sec:eval}
%%XM Here, we want to highlight the performance mismatches between the data loading and training, plus fine-grained breakdown analysis. 

%\begin{itemize}
    %\item 
    
    %\item Complex data augmentation: random crop, resize to (224,224), color jitter, random rotation, random horizontal flip, normalize.
%\end{itemize}

%We exercise a large number of data loading configurations, as shown in Table~\ref{tab:dataloadingconfig}.

In this section, we present experimental results for end-to-end training of various DNN models. Based on these results, we discuss the effect of the data loading pipeline on overall performance as well as how it results in underutilized resources throughout the training. 

\subsection{Experimental setup}
\noindent\textbf{Hardware and software setups:} Our experiments use AWS \\ \texttt{p3.16xlarge}, a high-end GPU instance, with 2.3GHz (base) %%XM and 2.7 GHz (turbo) 
Intel Xeon E5-2686 v4 processors, 488GB DRAM, 64 vCPUs, and 8 NVIDIA Tesla V100 GPUs (each having 5,120 CUDA Cores, 640 Tensor Cores, and 16GB GPU memory).  
%%XM \sam{Should we mention the vCPU count as well?} \yuxin{64 vCPUs}
OS and other software used include ubuntu18.04, CUDA 10.2, NVIDIA DALI 0.19.0, NVIDIA APEX 0.1, and PyTorch 1.4.0. 
Unless stated otherwise, we store training data via \texttt{ext4} on EBS~\cite{AWSEBS}, for durable, high-throughput %(14Gbps) 
instance-local block storage.
%%XM device attached to instances. The filesystem  is .
%It stores data on dedicated EBS space, \sam{what is EBS?} with 14Gbps bandwidth and 7500 IOPS through the ext4 file system.

\noindent\textbf{Preprocessing pipeline configurations:} 
Our data preprocessing pipeline contains the following operators: image decode, random crop, resize to (224,224), random horizontal flip, and normalize. 
We evaluate both raw data and record file preprocessing, each done by CPU-only or hybrid CPU-GPU processing, producing four combinations: \texttt{raw-cpu}, \texttt{record-cpu}, \texttt{raw-hybrid}, and \texttt{record-hybrid}. %\cheng{Need to explain how much work is done by GPU.}
As DALI is found to outperform built-in tools of major DL frameworks~\cite{Pieterluitjens2019DALIReport, DALI} (also confirmed by our tests), 
%%XM best performing data loading library, 
we adopt it for all our tests.
%%XM for the whole evaluation rather than t. 
Due to space limitations, we show only PyTorch results, with MXNet and TensorFlow behaving very similarly.  
%%XM \ping{what does last sentence mean? About figure3 (b) ?}
%%is PyTorch\cheng{We have try the other combinations, their performance numbers look similar}.

%%XM below commented by XM
\begin{comment}
\begin{table}[t!]
    \centering
    \footnotesize
    \begin{tabular}{c|c|c}
    Model&CPU-only&Hybrid\\\hline
    AlexNet & 512 & 512 \\
    ShuffleNet & 512 & 512 \\
    ResNet18& 512 & 512 \\
    ResNet50& 192 & 192 \\
    ResNet152& 128 & 128 \\
    \hline
    \end{tabular}
    \caption{Batch sizes for models
    \sam{Need to check the numbers; currently they are the same for CPU only and hybrid}}
    \label{tab:batchsize_config}
\end{table}
\end{comment}

\noindent\textbf{Models and data sets:}
We train five DNN models, namely, AlexNet~\cite{alexnet}, ShuffleNet~\cite{shufflenet}, ResNet18\cite{resnet}, ResNet50~\cite{resnet}, and ResNet152~\cite{resnet}, on the widely used ImageNet data set~\cite{ImageNet}. 
%\sam{Why reduced? Time? space? Give an explanation so that it does not look fabricated.}
We enable mixed-precision training (FP16 option on) to fully exploit GPU resources. 
For AlexNet, ShuffleNet, and ResNet18, we adopt a batch size of 512 as larger batches produce little performance improvement.
The other two models are much larger due to having more layers, so ResNet50 uses a batch size of 192 and ResNet152 uses 128, due to memory constraints. 
%%XM We evenly assign vCPUs for preparing data samples for 8 GPUs, each of which is bounded to 8 vCPU workers. 
%%XM Table~\ref{tab:batchsize_config} lists the batch sizes used, for obtaining the best training speed in each configuration. Compared with CPU-only data loading, hybrid loading uses smaller batch size for \tofill models so that \tofill. 
%%XM is reduced  hybrid loading is enabled, as the larger batch size resulted in GPU out-of-memory (OOM) errors in this case.  \yuxin{actually not}
%For the PyTorch built-in loader, ResNet50 uses a batch size of 128, while the other models use 512. \xm{Why?} 
%\sam{For DALI, we reduced the batch size for ReNnet18 and ShuffleNet to 384 and ResNet50 to 96 as the default batch size resulted in out-of-memory (OOM) errors.}
%\sout{For DALI, to avoid GPU out-of-memory errors, these three models use reduced batch sizes: 384 for resnet18 and shufflenet, 96 for resnet50.}
%%XM and shufflenet v2x1.1 is reduced to 384, 96, and 384, respectively.
%\xm{Here the models don't seem to be fully covered. Please carefully check and fill into the sentences.}

\noindent\textbf{Metrics:} We measure data preprocessing and model training throughput, in samples per second. 
%%XM To expose the negative impact of data loading, 
For reference, we also plot the ``ideal'' training performance, where 
%%XM the corresponding DNN model is constantly trained using 
training reuses a single batch preloaded into GPU memory. 
In addition, we measure both the CPU and GPU utilization levels.

\begin{figure}[t!]
    \centering
    \includegraphics[scale=0.38]{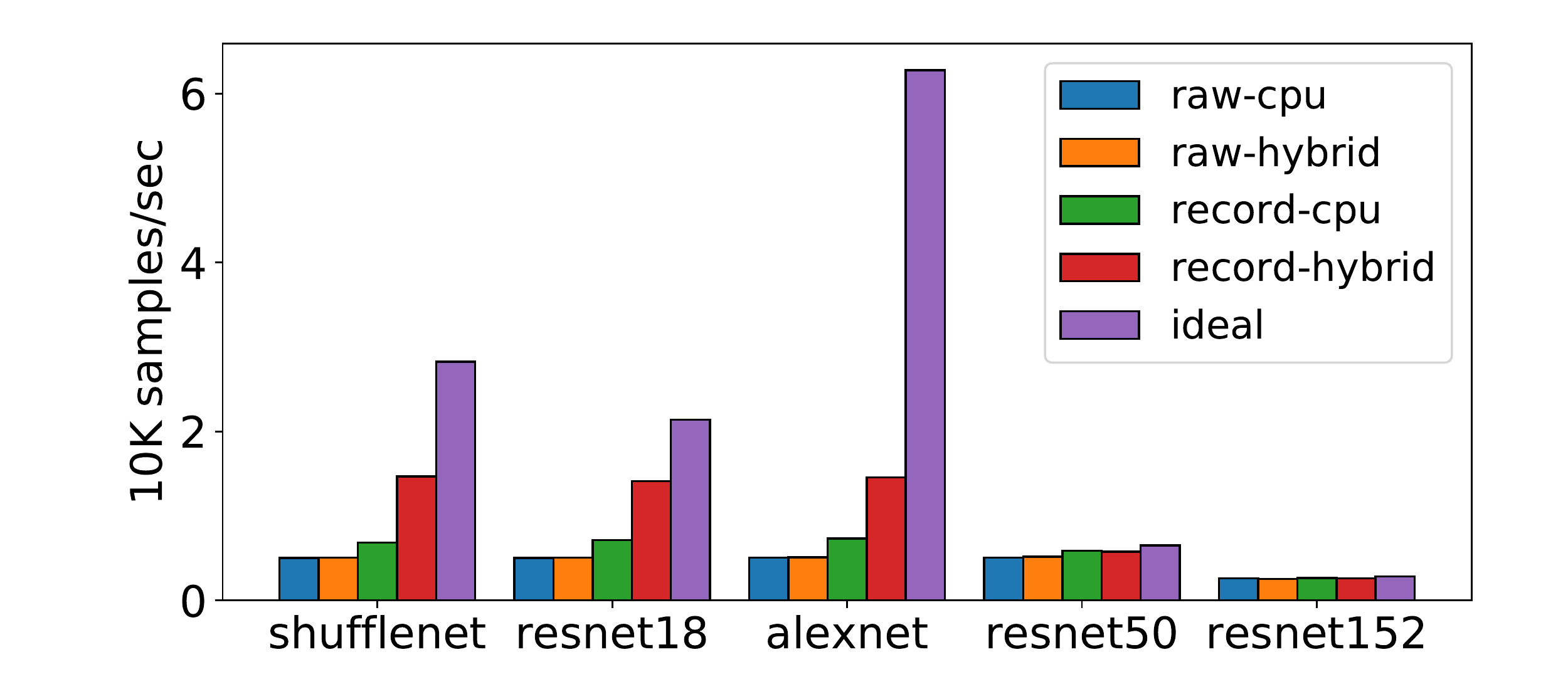}
    \caption{End-to-end training performance 
    %%XM with various loading configurations vs. the pure training performance. 
    %\cheng{There are two data points missing for resnet18 and resnet50.}
    %\sam{Use conventional names with appropriate capital letters for the NNs in the figures. Also raw-mixed, record-mixed have been renamed to *-hybrid.}
    } 
    \label{fig:fp16_simple_record0}
\end{figure}
\subsection{Overall performance}
%Figure~\ref{fig:fp16_simple_record0} summarizes the performance numbers. 
%\sam{To expose the negative impact of data loading, we additionally plot the ideal training performance, where the corresponding DNN model is constantly trained using a single batch that is preloaded into GPU memory.
%\textcolor{pink}{How about if we move this sentence to the last paragraph in 3.1?}
%}
%\sout{In order to discover the mismatch between data loading and GPU capabilities, we additionally plot the ideal training performance, where we make GPU train the corresponding DNN model constantly using a single batch that is preloaded into GPU memory. This setting enables us to completely remove the negative impacts of data loading.}

Figure~\ref{fig:fp16_simple_record0} gives overall training performance across 
%%XM five exercised DNN models combined with four different 
preprocessing methods. 
First, most models see improvement when using record files and especially when adopting hybrid data preprocessing.
The former benefits from a higher ratio of sequential I/O, which is advantageous even on NVMe SSDs. 
The latter allows powerful GPUs to ease the data preprocessing bottleneck on CPUs and achieve more balanced system utilization. 

However, we observe that for three of the five models, their best end-to-end training throughput is significantly lower than the ideal one. This demonstrates the performance impact of the 
%\textcolor{green}{
data preprocessing pipeline
%}
and GPU under-utilization. 
%\textcolor{green}{
That is, the data preprocessing pipeline is not able to fill the GPUs fast enough such that the GPUs are being underutilized, which is validated in the following section.
%}
In particular, for AlexNet, its peak performance under \texttt{record-hybrid} is only 23\% of the ideal case.
Considering these three fast data consumers where hybrid data preprocessing helps the most (exemplified by the large jump from green bar to red bar), as the GPUs otherwise starved for data is used to offload data preparation from CPUs, we observe 98\% to 114\% throughput improvement for them compared to \texttt{record-CPU}.
Interestingly, even for these three models, hybrid processing does not help when raw files are used, as random I/O dominates loading and CPUs alone can match the I/O speed. 
%%XM Given that the performance of accelerators are expected to improve~\cite{PattersonGoldenAge}, this mismatch would offset the advance of those accelerators. 
ResNet50 and Resnet152, in contrast, are much more computation-intensive in training (with much lower ideal training throughput than the other models), rendering data loading less influential.  
%%XM This implies that, while the 8 GPUs can process more data samples, they are sitting idle waiting for the data loading pipeline to supply the data.\ping{it`s not right, for gpus speed a lot of time in doing data augmentation in alexnet-record-hybrid} 
%%XM \sam{We now present a detailed analysis to better understand where this mismatch is coming from.???}\cheng{We will do this in the resource utilization part.}

\begin{figure}[t!]
    \centering
    %\subfloat[CPU-Only]{
    \includegraphics[scale=0.23]{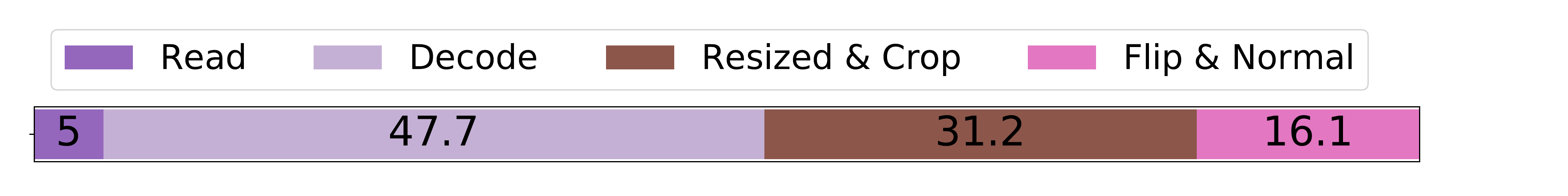}
    %}
  %  \\
   % \subfloat[Hybrid CPU side]{
   % \includegraphics[scale=0.35]{images/decoding_cost_on_cpu.png}
  %  }
    \caption{100\% stacked bar graph showing breakdown latency of major steps for preprocessing a single image on CPU, which takes 14.26 ms in total (Numbers are percentages.)
    %\ping{ should we mention for one image}
    %\cheng{Please remove complex}
    }
    \label{fig:timecostCPU}
\end{figure}

\begin{figure*}[t!]
\centering
\subfloat[GPU]{
    \includegraphics[scale=0.29]{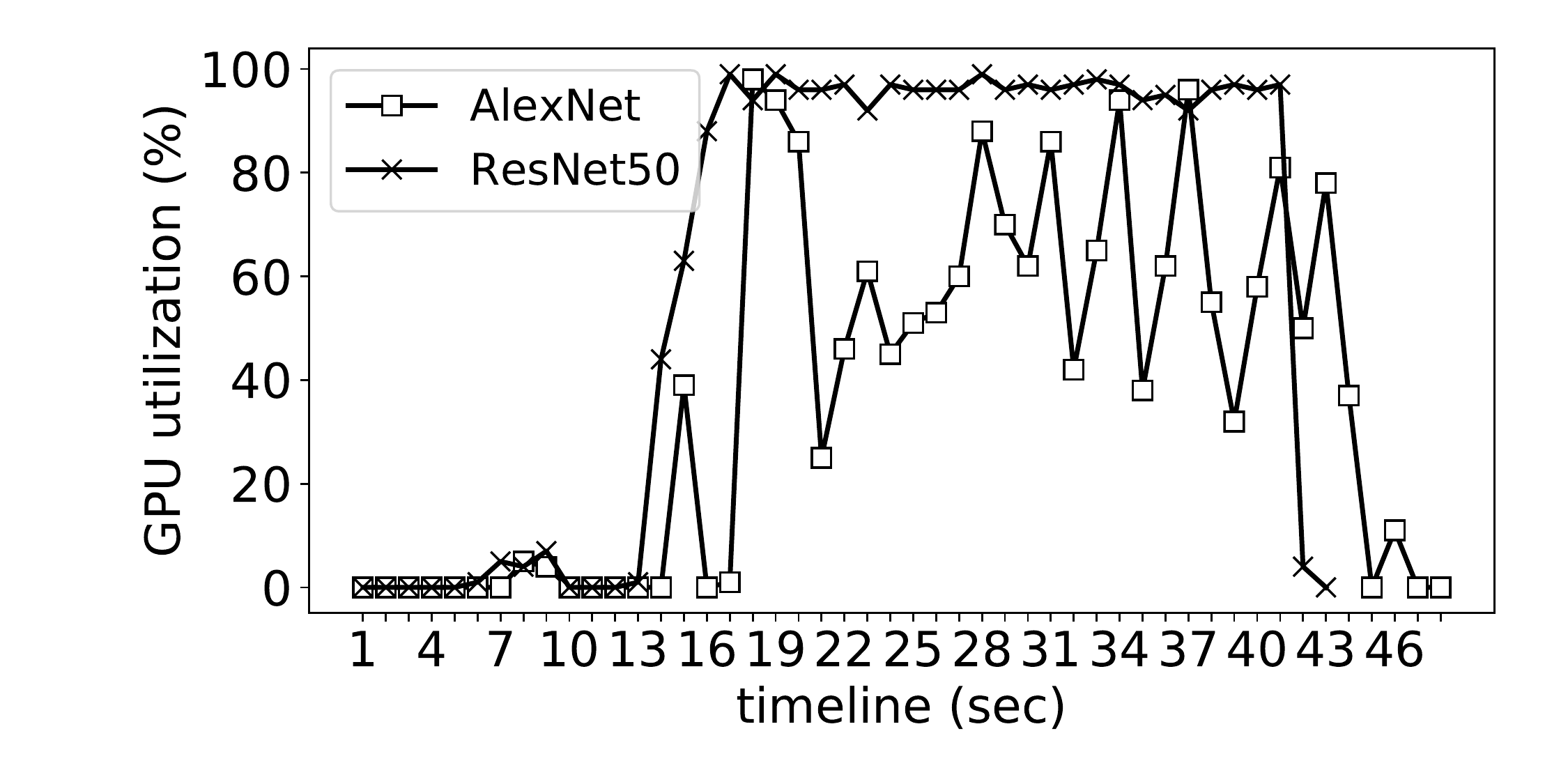}
    \label{fig:res_ut_gpu}
}
\subfloat[CPU]{
    \includegraphics[scale=0.29]{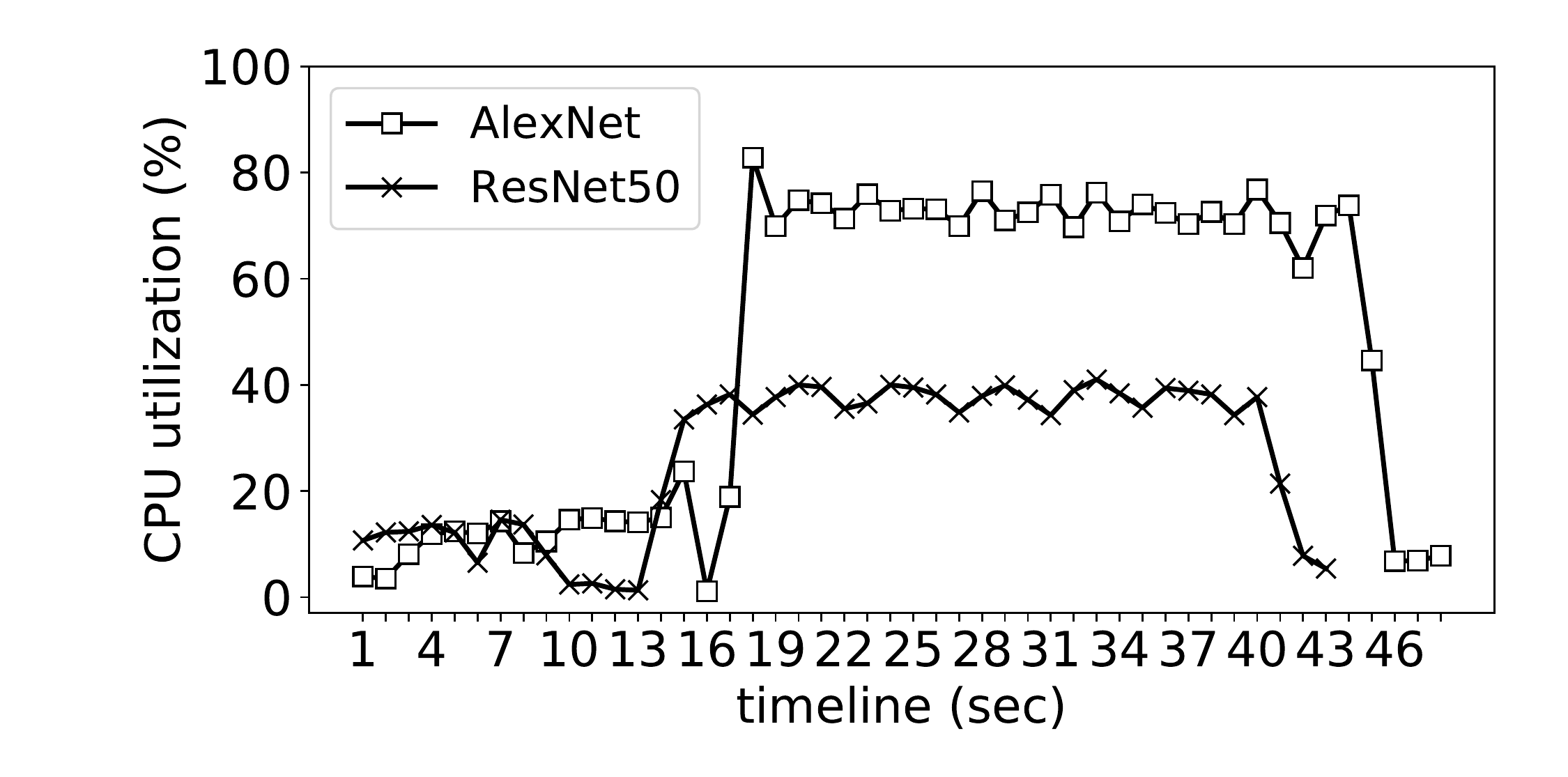}
    \label{fig:res_ut_cpu}
}
\subfloat[I/O]{
\includegraphics[scale=0.29]{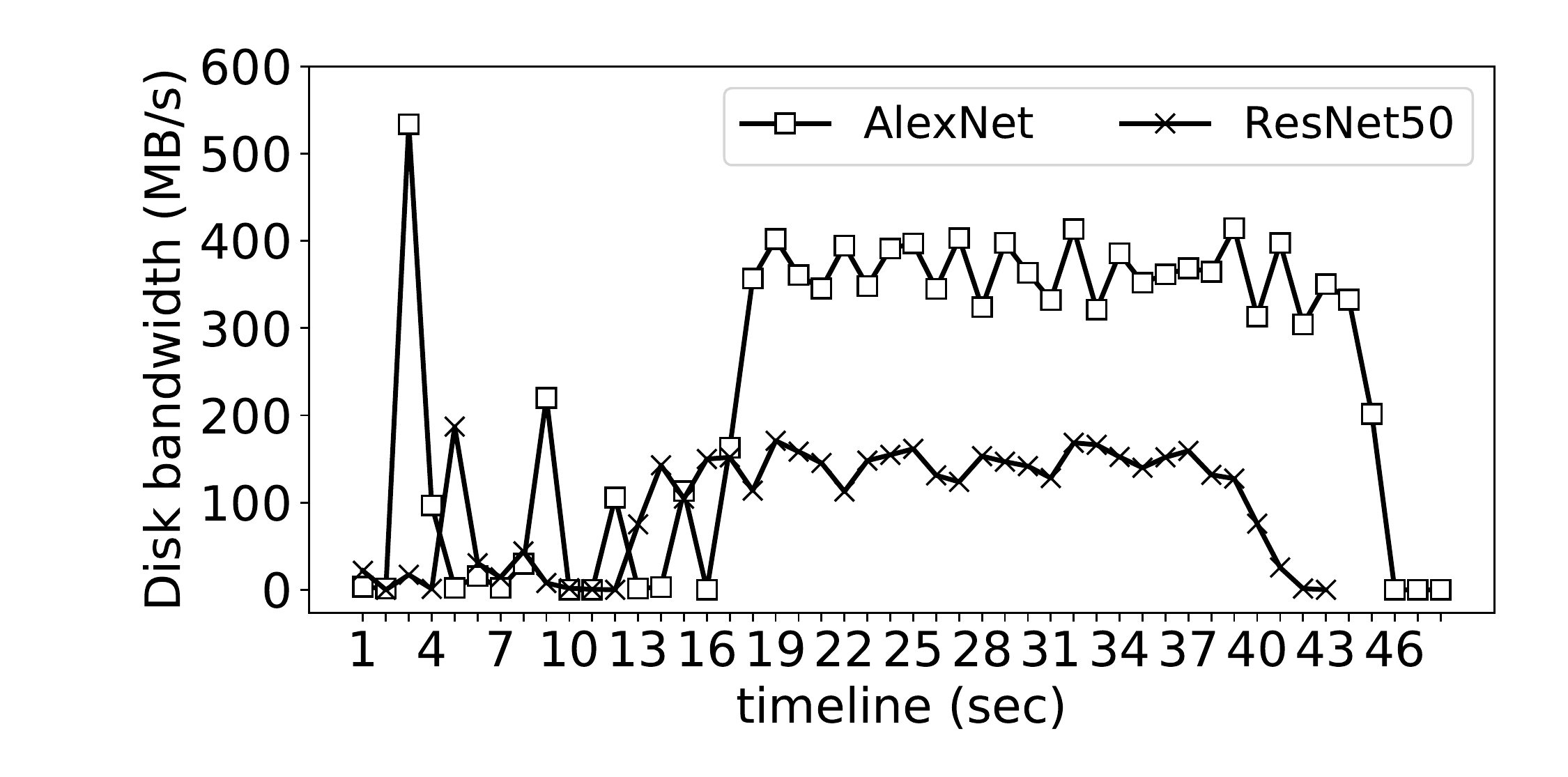}
    \label{fig:res_ut_io}
}
\caption{Resource stats of AlexNet and ResNet50 corresponding to the best performed \texttt{record-hybrid} experiments in Figure~\ref{fig:fp16_simple_record0}.
%\sam{
%1) AlexNet and ResNet legends are flipped. 
%2)ResNet $=>$ ResNet50
%, 3) add unit to x-axis, I think minutes %4) (a)(b)(c) all have double parenthesis 
%5) y-axis of (c) should be bandwidth (or throughput) not utilization? 6) (a) y-axis is 80\%? Not 100\% 
%7) (a) CPU utilization?? NOT GPU utilization? 8) (b) y-axis GPU utilization?? NOT CPU ?? 9) OR MAYBE (a) and (b) are flipped?
%}
    }
\label{fig:res_utilization}
\end{figure*}
%Figure~\ref{fig:timecostCPU} gives \tofill. 
%\color{green}
To further investigate why offloading portions of preprocessing to the GPUs significantly improves training performance, here, we report the latency breakdown of different steps along the data loading pipeline. 
In so doing, we do not enable training, instead, only focus on running the data preprocessing pipeline and exercising the performance of its individual operators. 
We plot the times spent in different data augmentation operators in Figure~\ref{fig:timecostCPU}.
Note that we are reporting breakdowns when only the CPU is being used for data preprocessing.
The results highlight that all data preprocessing operators consume nearly 95\% of the whole pipeline cost. Among these operators, image decoding is most time-consuming, which accounts for 47.7\% of the total time. 
When \texttt{record-hybrid} is used in our experiments, data preprocessing splits the operators such that most of the decoding and beyond are performed by the GPU, reducing the overall preprocessing time.

Note that the performance benefits of \texttt{record-hybrid}
%%XM combining record files with CPU-GPU co-processing 
come at a cost. 
First, it requires extra work offline to produce record files.
%%XM resulting in time, computation and storage costs. 
%\color{green}
Second, effective pipeline design requires concurrent data preprocessing and model training on the GPUs.  
%%results in sharing of the limited GPU resources with DNN model training.
This comes with two consequences.
One is that there is a potential cause for OOM errors, which will require manual intervention to avoid.
The other is that, as we show later in Figure~\ref{fig:reconfigure}, training throughput may be negatively affected due to the sharing of the GPUs.
\subsection{Resource Utilization}
%%XM Here we understand the causes of bottlenecks and the efficiency of resource allocation by zooming in the resource utilization statistics of AlexNet and ResNet50, corresponding to the yellow bars in Figure~\ref{fig:fp16_simple_record0}. 
Figure~\ref{fig:res_utilization} plots the utilization level of different resources during the training process under \texttt{record-hybrid} for AlexNet and ResNet50, each, respectively, representing a fast and slow data consumer. It is worth noting that the roughly first one-third of each curve corresponds to the initialization of training.   
For ResNet50, GPUs are nearly saturated for the last two-thirds of the entire process while CPU utilization is at a much lower 38\% and the I/O bandwidth-consuming only 147MB/s, also much lower than its maximum capacity.
%\xm{What are the GPUs doing in the first 1/3 and why isn't training started earlier?}
%\sam{Note that the first one-third of the process is for initial preprocessing of the data.???}
%%XM , only \tofill yuxin{maximum not clear, assume 400 according to alexnet case, 37}\% of its maximum. 
%\color{green}
We observe that ResNet50 is GPU-bound where training data consumption is slow enough to hide I/O and data preprocessing.
Thus, we find that CPU and storage devices are being underutilized.
We also see that even though GPU resources are precious, DALI is making use of these resources for data loading, which is an unwise choice of use.
%\color{black}

%% This is because for this model training is already highly computation intensive such that I/O and data loading can be mostly hidden behind the computation. In such case, there is no need to (1) offload preprocessing to GPUs to contend for already constrained GPU resources; (2) to allocate a large number of CPU cores for its data loading pipeline that the demands; and (3) to upgrade the storage hardware with higher throughput. 

%\color{green}
AlexNet, on the other hand, is a fast data consumer making quick use of the loaded data and then waiting for more data to arrive.
This is reflected in the fluctuation of GPU utilization.
Overall, the GPU is utilized under 50\%.
In contrast, we see that the CPU utilization much higher than that of ResNet50 (Figure~\ref{fig:res_ut_cpu}) and that the I/O bandwidth is also considerably higher (Figure~\ref{fig:res_ut_io}).
We see that \texttt{record-hybrid} is helpful in this case as it helps improve the utilization of the GPUs.
However, we see that the CPUs are not enough to satiate the data needs of the GPU and that the more CPUs need to be allocated to the data preprocessing pipeline.

\section{Effect of Resource Balancing}%% on End-to-End Training}
\label{sect:presolution}
\begin{table}[!t]
    \centering
    %\small
    \caption{VM instances commonly used for DNN model training on AWS EC2~\cite{EC2GPUInstances} (top), %Azure~\cite{AzureGPUInstances}, 
    and Google Cloud~\cite{GoogleGPUInstances} (bottom). All GPUs are V100. 
    %%XM \textbf{varied} indicates that one can configure the size, while \textbf{choices} indicates that one can only choose from a limited number of options. Price is measured as \$/hour.
    }
        \begin{tabular}{c||c|c|c|c}
          Type & \#GPU & \#vCPU & I/O & \$/h \\\hline \hline
        p3.2xlarge & 1 & $\leq8$ & Configurable & $< 3.06$\\
         %& p3.8xlarge & 4 V100 & 32 & 244 & 7000 & 12.24\\
         p3.16xlarge& 8 & $\leq64$ & Configurable & $<24.48$\\
         p3dn.24xlarge & 8 & $\leq96$ & Configurable & $<31.21$ \\\hline
         %\multirow{4}{*}{Azure} & ND6rs & 1 P40 & 6 & 200 & 2.07\\
         %& ND12s & 2 P40 & 12 & 224 & 400 & 4.14 \\
         %& ND24s & 4 P40 & 24 & 800 & 8.28 \\
          %& ND40rs\_v2 & 8 V100 & 40 & 800 & 22.032 \\\hline
         V100-1 & 1 & $\leq12$ & Options & $<3.22$ \\
          %&V100-2& 2 V100 & 1 - 24& 1 - 156 & -& 6.45 \\
          V100-4 & 4 & $\leq48$ & Options & $<12.90$ \\
          V100-8 & 8 & $\leq96$ & Options & $<25.80$\\\hline
    \end{tabular}
\label{tab:cloudsurvey}
\end{table}

Public cloud providers are beginning to offer more flexible instance configurations. 
Table~\ref{tab:cloudsurvey} lists available GPU VM instances on Amazon EC2 and Google Cloud. 
%%XM We've noticed that both AWS EC2 and Google Cloud enable flexible node configurations to some extent.
With a given instance type associated with fixed GPU resources, both platforms allow users to specify the desired number of vCPUs within a preset range.
%%XM For instance, Google Cloud allows to change the number of vCPU in a preset range. 
EC2 further enables I/O throughput (proportional to persistent storage capacity requested), while Google Cloud provide a limited number of storage choices. 
Along with such flexible configurations come fined-grained charging policies.
%%XM Besides the configurations, we also find that public cloud providers charge customers in a more fine-grained manner than what we expected before. For instance, 
\Eg, on the listed Google Cloud instances, GPU and vCPU cost 2.48 and 0.033\$/hour, respectively, while memory costs 0.0044\$/GB-hour.
%static configurations and prices. The conclusion is that both EC2 and Azure offer static configurations\yuxin{ec2 allows people to decrease the vCPU number}, while Google starts to allow people to create instances with a configurable number of vCPU and Memory. 
%%XM The new trends taking effects in public clouds encourage us to explore if we can address performance bottlenecks while achieving cost-efficiency by configuring the number of vCPU and I/O for a fixed number of GPUs.
%%XM This creates an unique opportunity to adjust the resource allocation for each model. For instance, we can vary the number of CPU workers for different models so that we can eliminate the data preprocessing bottleneck while completing the training jobs in a cost-efficient manner. 

%\sam{I think this is another important part that which we have not been discussing.
%As this is a technical (short, position) paper, there needs to be a technical contribution in terms of resolving (some or part of) the problems observed.
%This may be preliminary, but there needs to be some type of solution.}

%Here, we present a simple but working solution and the preliminary results.

%%XM \subsection{New trends in cloud resource allocation}
%Unfortunately, with today's cloud infrastructure where static node configurations are mostly offered, it is impossible to meet this diverse and fine-grained resource allocation requirements. 

%%XM \subsection{Benefits from careful configurations}

\begin{figure}[t!]
    \centering
    %\subfloat[Adjust I/O for AlexNet]{\includegraphics[scale=0.35]{images/resnet-cpu-mixed.png}
    %\label{fig:reconfig_alexnet_io}
    %}
    %\\
    \subfloat[Adjust \#vCPU for AlexNet]{\includegraphics[scale=0.255]{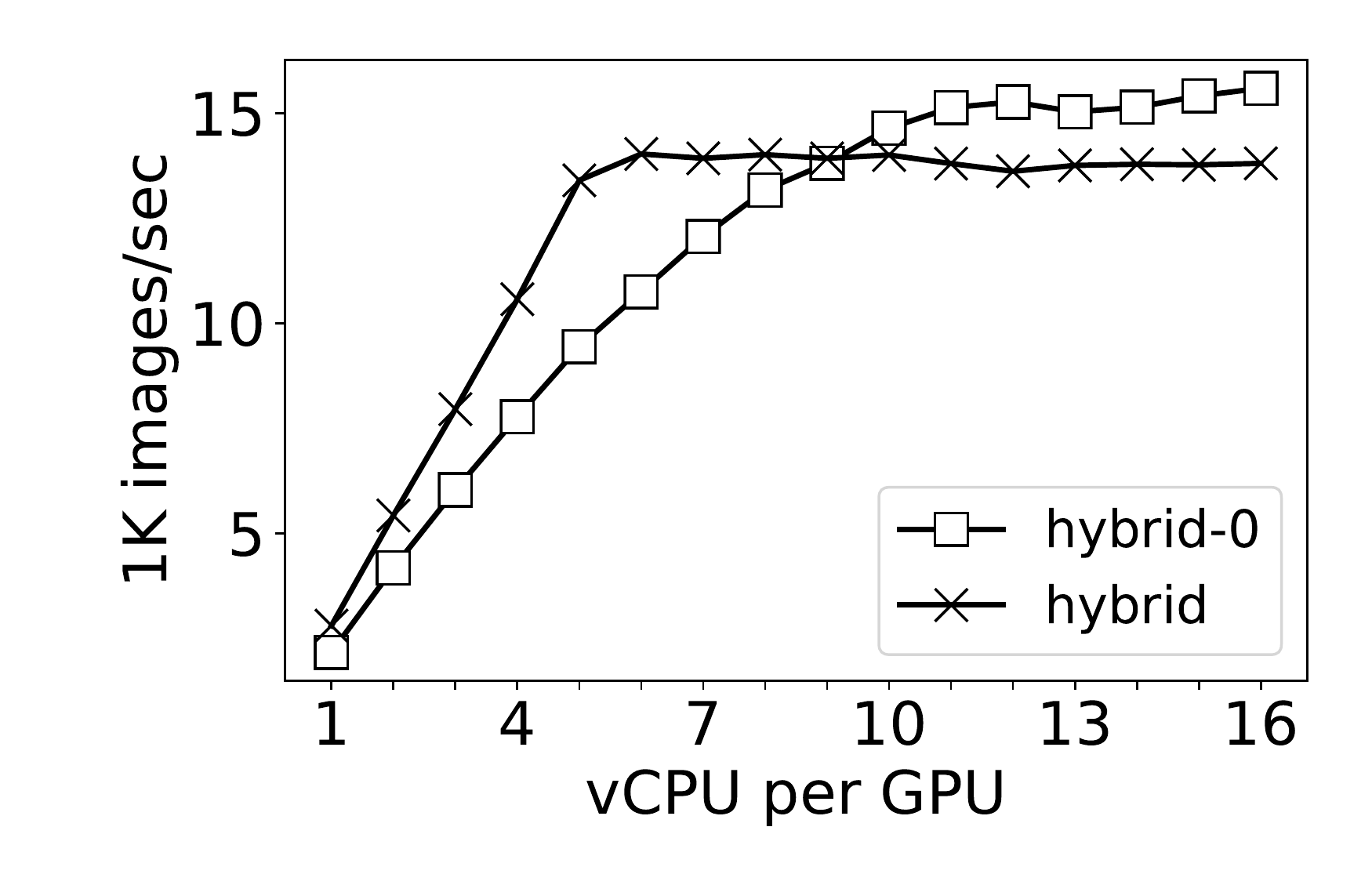}
    \label{fig:reconfig_alexnet_cpu}
    }
    \subfloat[Adjust \#vCPU for ResNet50]{\includegraphics[scale=0.255]{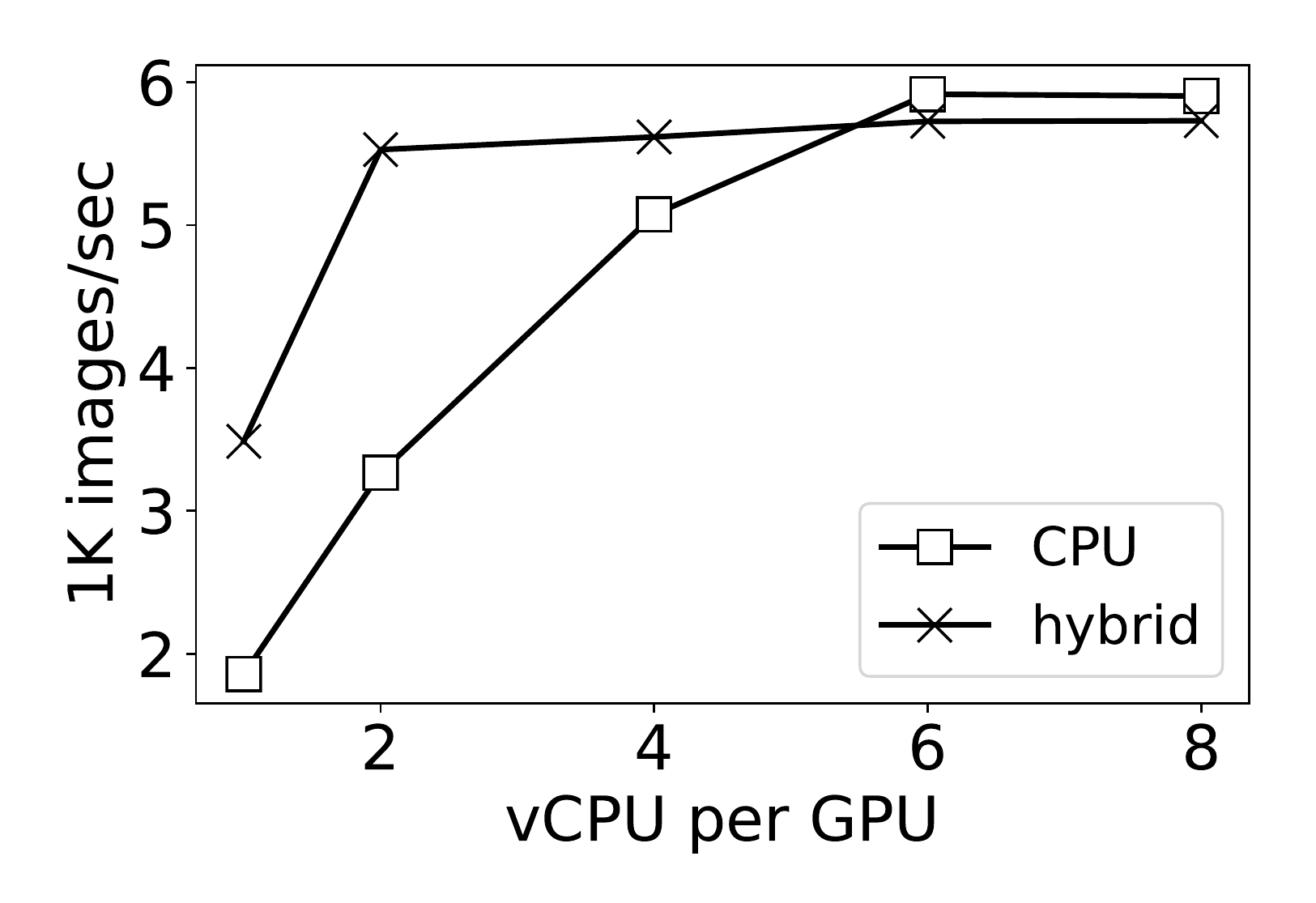}
    \label{fig:reconfig_resnet_cpu}
    }
    \caption{End-to-end training performance of AlexNet and ResNet50 with CPU resources adjusted for GPUs
    } 
    \label{fig:reconfigure}
\end{figure}

While such flexible configurations provide consumers with better choices, selecting the best configuration for a DNN model is not an easy task.
Our eventual goal is to develop a tool that will propose model-specific, fine-grained resource configurations for a model training workflow while maintaining high throughput performance. 
Our findings described in the previous section show that such a tool will be valuable.

In this section, we present some results that may be possible once this tool is developed.
In particular, we investigate the performance implications of adjusting the number of vCPUs. % \cb{on the aforementioned 8-GPU EC2 instance}.
Again, in the interest of space, we focus on the two representative DNN models evaluated earlier, AlexNet and ResNet50. 
%%XM In the evaluation section, as shown in Figure~\ref{fig:res_ut_io}, we observe that the I/O could be a bottleneck, as GPUs need more data samples for training AlexNet. Therefore, our first attempt is to 
%First, we adjust the capacities of I/O devices for AlexNet, shown to be loading-bound in Figure~\ref{fig:res_ut_io},
%with three types of EBS devices (providing \tofill, \tofill, and \tofill IOPS, respectively). 
%Figure~\ref{fig:reconfig_alexnet_io} shows the end-to-end training performance.
%%XM of AlexNet with different I/O settings. 
%Evidently, faster storage device brings \tofill\% enhancement.
%Note that the accelerated data loading in turn requires more vCPUs to saturate GPUs. 
%For instance, with the fastest device, we need \tofill vCPUs, \tofill\% more than
%called by the slowest one.
%%Here, we examine CPU-GPU balancing in data preprocessing in two set of experiments. 

First, consider the preprocessing-bound AlexNet, where we fix the number of GPUs to 4 and adjust the number of vCPUs. 
With all 64 vCPUs available for the 4 GPUs, we investigate one additional loading method (aside from the \texttt{hybrid} mode evaluated earlier), which performs decoding on CPU and the rest of preprocessing on GPU, denoted as \texttt{hybrid-0}. 
This mode uses more CPU resources while relieving the more expensive GPU resources to training. Recall that in a \texttt{hybrid} image decoding is jointly performed by CPU and GPU. 
%%XM Obviously, the second data loading requires more CPU resources than the first. However, the second one also has an advantage of not stealing much resources from GPUs. Therefore, as shown in 
Figure~\ref{fig:reconfig_alexnet_cpu} shows that, while \texttt{hybrid} saturates training at 24 vCPUs (6 per GPU), \texttt{hybrid-0} delays its saturation point to 44 vCPUs (11 per GPU).
This results in a 7.86\% increase in training speed as depicted by a higher line beyond 11 vCPUs in the figure. 

Now consider the training-bound ResNet50, where the GPU is the clear bottleneck. 
Thus, we make use of all 8 GPUs for these experiments.
%%XM As shown in Figure~\ref{fig:res_ut_gpu}, GPU is the bottleneck for training ResNet50. Therefore, 
Here we can explore reducing vCPU allocation without performance degradation or improving performance by re-configuring the CPU-GPU co-processing tasks. 
Figure~\ref{fig:reconfig_resnet_cpu} confirms that
%%XM the end-to-end training performance of Resnet50, where we fix the number of GPUs to 8, while we adjust the number of vCPU for loading data. The conclusion is that for resnet50, across all configurations, 
the number of required vCPUs here is much lower than available in such instances. For instance, under the \texttt{hybrid} mode, 16 vCPUs (2 per GPU) can adequately feed the GPUs. 
%%XM Since at this point GPU is the bottleneck, it makes no sense to offload parts of preprocessing to GPUs. 
To further reserve GPUs for the time-consuming training, we push DALI optimization back to using solely CPU for data preparation, \ie, the \texttt{cpu} mode, denoted \texttt{CPU} in the figure. 
Now we need 48 vCPUs (6 per GPU) to saturate the training pipeline, paying more CPU costs, for a training speed improvement of 3.03\%. 
We further observe (but not shown here) that moving from ResNet50 to ResNet152, with deeper layers, the training pipeline becomes even more GPU-bound, reducing the vCPU requirement to 8 vCPUs (1 per GPU).
Note that in both the AlexNet and ResNet50 examples, we observe that employing GPUs for the data preprocessing pipeline, though helpful for the pipeline, results in reduced throughput.
%%XM are enough to meet the GPU data loading demands.

%\paragraph{Operator fusion to make the latter operator reuse the intermediate results produced by the former one so that parts of computation can be avoided}

%\paragraph{Parallelism reconfiguration. This may be only applicable for Tensorflow.}

\begin{figure}[!t]
    \centering
    \includegraphics[scale=0.35]{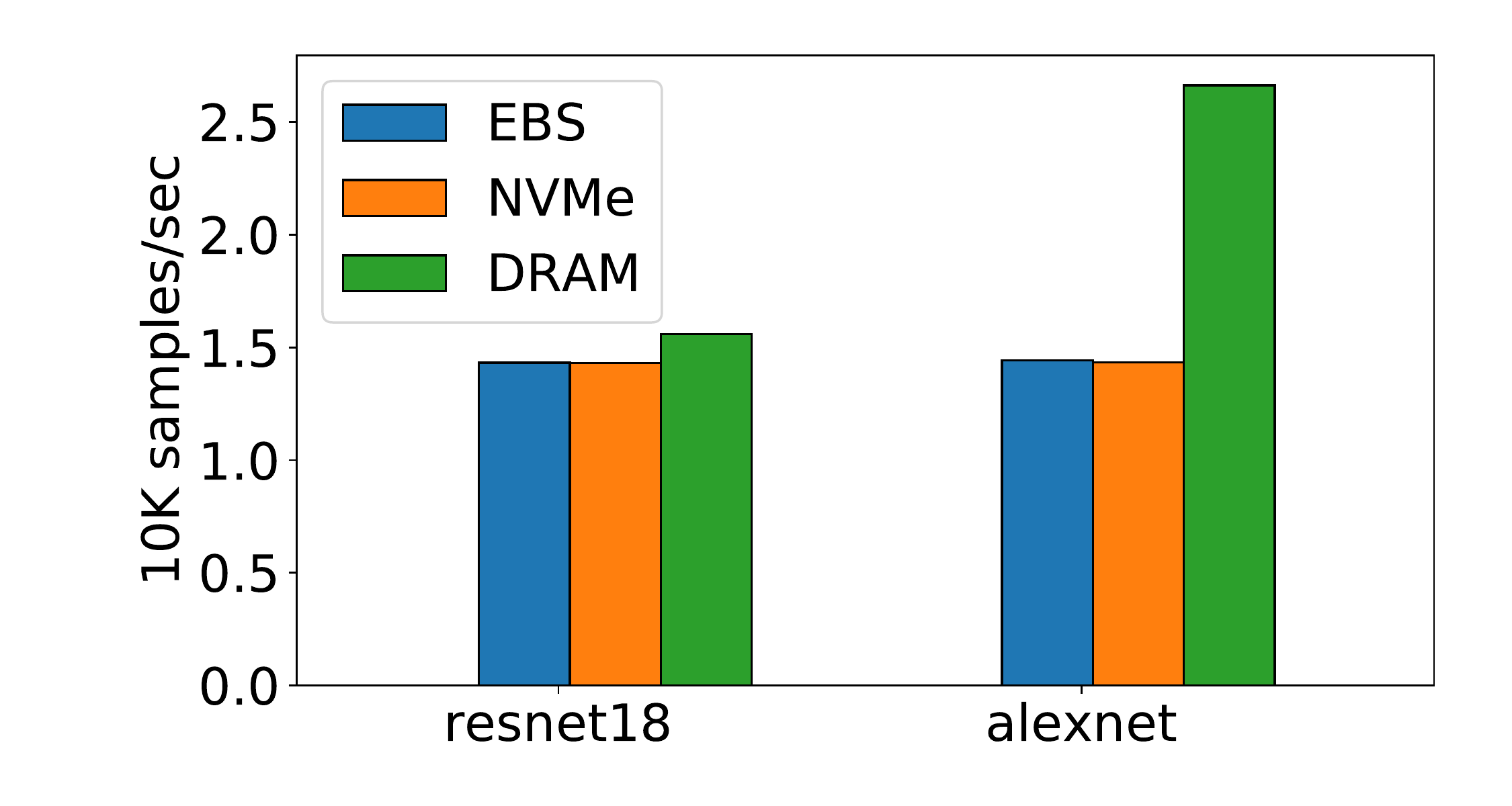}
    \caption{End-to-end training performance of ResNet18 and AlexNet, loading data samples from different storage options}
    %in AWS p3dn.24xlarge with ebs of 7500 IOPS, by using 4 v100 GPU card and 48 vCPU cores.
    \label{fig:IO}
\end{figure}

%\subsection{Impact of I/O}
Finally, we shift our attention to exploring the performance implications of using different storage media for hosting training data. 
For this purpose, we deploy AlexNet and ResNet18 on an EC2 p3dn.24xlarge instance, and train them with 4 V100 GPUs, each assisted by 12 vCPU cores. 
We vary the storage devices, testing EBS (up to 7500 IOPS), two NVMe SSDs (default setting by EC2), and DRAM.
Figure~\ref{fig:IO} reports the end-to-end training performance.

We observe that loading data samples directly from EBS and NVMe SSDs (corresponding to \ding{175} in Figure~\ref{fig:pipeline}) achieves almost the same training performance for the two models. 
This is because EBS is a highly optimized storage option exploring SSDs and it offers similar I/O bandwidths as the attached NVMe SSDs. Interestingly, when loading data samples from DRAM, the training performance of ResNet18 has been just improved by 8.8\%. 
This is a direct consequence of the fact that ResNet18, like ResNet50, is computation-intensive and the bandwidths of EBS and NVMe SSDs are sufficient to meet the needs of the training phase in GPU. 
In contrast, using DRAM leads to a 1.84$\times$ speedup of the AlexNet training performance against the other two storage options. 
This is because the computation of AlexNet is much faster than ResNet18, and it demands larger I/O bandwidth beyond the capacity of both EBS and NVMe SSDs. 
Therefore, the performance impact of storage configuration is model dependent. 
For certain models, investing in high-bandwidth storage may significantly improve the overall cost-effectiveness of DNN training.

%%XM In summary, in addition to the CPU-GPU configuration, it is also necessary to consider the balance between I/O bandwidths and target workloads.

%\input{related}
%\input{appendix}
%\clearpage
%\clearpage
% \section*{Discussions}
% \label{sect:discussions}

\section{Conclusion and Future Directions}
It is commonly accepted that DNN training requires considerable resources in terms of  GPU/accelerator and network infrastructure. 
While studies to expedite training  through better management of these resources have been the focus of many studies, we contend that there is another aspect that is as important, which, to date, has been largely neglected.
Specifically, data has to be fed to the accelerators such that the accelerators can be fully utilized.

In this work, we examine the data preparation process of widely used deep learning model training frameworks. 
We have found that different frameworks have highly similar data loading and preprocessing pipelines, while the relative overhead of data preparation to that of in-memory model training varies significantly across popular DNN models. Based on our performance results, our contention is that this pipeline is currently inefficient, often leading to considerable resource waste, especially the under-utilization of expensive GPU resources.
We further verify the effectiveness in hybrid CPU-GPU data preparation and propose model-specific, fine-grained cloud instance configuration for efficient and cost-effective model training. 

In future work, we will extend our performance analysis to both NLP and video processing models. 
Following that, we plan to build a tool that enables automatic resource configuration to adapt to the demand for diverse deep learning training workloads.

\bibliographystyle{ACM-Reference-Format}
\bibliography{main}

\end{document}